\documentclass{acm_proc_article-sp}
\usepackage{graphicx}
\graphicspath{ {images/} }
\usepackage{bbm}
\usepackage{tikz}
\usetikzlibrary{bayesnet}
\begin{document}

\title{A Bayesian Ensemble for Unsupervised Anomaly Detection}

\numberofauthors{2} 
\author{
\alignauthor
Edward Yu\\
       \affaddr{Columbia University}\\
       \affaddr{1130 Amsterdam Ave}\\
       \affaddr{New York, NY 10027}\\
       \email{edward.yu@columbia.edu}
\alignauthor
Parth Parekh\\
       \affaddr{Facebook Data Engineering}\\
       \affaddr{1 Hacker Way}\\
       \affaddr{Menlo Park, CA 94025}\\
       \email{parthparekh@fb.com}
}
\date{21 July 2017}

\maketitle
\begin{abstract}
Methods for unsupervised anomaly detection suffer from the fact that the data is unlabeled, making it difficult to assess the optimality of detection algorithms. Ensemble learning has shown exceptional results in classification and clustering problems, but has not seen as much research in the context of outlier detection. Existing methods focus on combining output scores of individual detectors, but this leads to outputs that are not easily interpretable. In this paper, we introduce a theoretical foundation for combining individual detectors with Bayesian classifier combination. Not only are posterior distributions easily interpreted as the probability distribution of anomalies, but bias, variance, and individual error rates of detectors are all easily obtained. Performance on real-world datasets shows high accuracy across varied types of time series data.
\end{abstract}

\category{I.5.4}{Pattern Recognition}{Applications}

\terms{Theory}

\keywords{anomaly detection, outlier detection, ensemble methods} 

\section{Introduction}
Anomaly detection or outlier detection refers to identifying certain subsets of data which are inconsistent with the remainder of the data, so much so that they arouse suspicion that they were generated by a different underlying process. Anomaly detection algorithms have seen widespread use in a number of settings, including credit card fraud detection, network intrusion monitoring, and military surveillance systems. As such, numerous detection schemes have been tried: from statistical methods in the 19th century to more recent machine learning techniques \cite{Chandola}. As Chandola et. al pointed out, each detection scheme makes different assumptions about the underlying data; each algorithm has to precisely define an anomaly in order to detect it~\cite{Chandola}. This may involve, for example, assuming that normal data clusters around a mean and defining distance from the nearest neighbors as a measure of anomalousness. Other algorithms may assume periodicity in the data and use phase shifts from that periodicity as a indication of abnormal behavior. These assumptions lead to heterogeneous behavior and accuracy when used across different types of datasets. 

One attempt to improve robustness and accuracy is in the form of ensemble methods. An ensemble uses a collection of individual learning algorithms to produce a consensus. The hope is that although one or two learners may be off base, the majority will be able to produce the correct decision. 

There are several challenges with using ensembles, however. 
\begin{enumerate}
  \item The data that is provided to any unsupervised learning algorithm is unlabeled, meaning that there is no a priori knowledge about which points are outliers and which points are inliers. A simple majority voting scheme has been shown to be suboptimal \cite{Arora}, but the optimal voting strategy involves knowing the true labels of the data and the error rates of the individual algorithms. 
  \item Each individual learning algorithm must make decisions independently and have uncorrelated errors, else the ensemble could produce worse results than the individual learners \cite{Dietterich}. 
\end{enumerate}

Our contributions are:

\begin{enumerate}
    \item We offer a method for constructing ensembles specific to anomaly detection. This ensemble is fully unsupervised and does not require labeled training data, which in most practical situations is hard to obtain.
    \item For the first time, we adopt Bayesian classifier combination to anomaly detection. Unlike previous ensemble approaches to anomaly detection, all data is modeled as probability distributions. This is more rigorous than simply averaging the output scores of individual detectors, and the results are more theoretically sound. A wealth of previously incalculable data is now easily extracted. For example, not only does the model measure whether a point is an anomaly or not, but we can also measure the variance of this estimate, as we have the entire posterior distribution. As another example, the model also estimates individual error-rates of each individual detector. Bias and variance of each detector can also be calculated. These new metrics can serve as a foundation for further improvements. 
    \item We provide analysis showing that the ensemble is robust across a wide variety of time series. Further, we show that the ensemble is robust even in the case where some individual detectors are inaccurate, a distinguishing feature from other ensembles.
\end{enumerate}
The rest of the paper is organized as follows. Section 2 gives an overview of related work. Section 3 details several anomaly detection algorithms used to construct the ensemble and an analysis of the assumptions that they make of the underlying data. Section 4 describes a Bayesian updating meta-algorithm that jointly estimates true labels and individual error rates. Section 5 presents the results of the algorithm as tested on real-world data. Section 6 concludes. 

\section{Related Work}
Ensemble methods for anomaly detection have seen relatively little attention in the literature. Zimek et. al provide an overview of current research and challenges \cite{Zimek}. Many papers investigate model averaging: taking the output scores from individual detectors and combining them in some way (average, max, min) to produce a final anomaly score \cite{Aggarwal} \cite{Chiang} \cite{Gao}. There are also a variety of more complicated approaches at creating consensus \cite{Strehl} \cite{Gao2} \cite{Lazarevic}. Chiang and Ye suggest that it is surprisingly hard to beat a simple average of the outputs of individual detectors \cite{Chiang}. Although model averaging performs well in certain cases, there are several features to be desired. Averaging scores produces a number that is hard to interpret, as it definitively not the probability that a data point is an anomaly. An average of output scores is also sensitive to inaccurate detectors, especially when the inaccurate score deviates far from the average. 

There is also some work in the direction of selection algorithms: choosing which detectors to include in an ensemble to mitigate the above problem \cite{Rayana} \cite{Schubert}.

Combining the opinions of experts is a problem that has surfaced in several other fields, and is the predecessor to our work. In their seminal work, Dawid and Skene~\cite{Dawid} describe a situation in which independent anaesthetists ask a patient questions in order to obtain a diagnosis, but the patient may give different replies to the same question. The goal is estimate the individual error rates of each doctor as well as the true response of the patient. They propose the following expectation maximization algorithm to uncover the true response. After initialization, iterate until convergence: 
\begin{enumerate}
  \item Estimate true responses with maximum likelihood estimates of individual observer error rates. 
  \item Using the estimates of true responses, estimate the individual observer error rates. 
\end{enumerate}

There are several limitations to this algorithm. First, EM algorithms are only guaranteed to converge to local maximums and not global maximums. Second, the algorithm does not perform so well in practice; it was found that the error rates are several orders of magnitude higher than Bayesian models \cite{Platanios}.  

Kim and Ghahramani \cite{Kim} and Platanios et al.\cite{Platanios} extend this work with a Bayesian model. They propose a Bayesian framework which is able to jointly estimate the true response as well as a confusion matrix for each individual detector. The confusion matrix will contain the probability of Type I and Type II errors.

We adapt this work in the context of anomaly detection. In the case of general classifiers, each classifier labels a data point. When the classifiers are detectors, not only are labels for data points provided, but also an ``anomaly score" indicating the strength of belief is outputted. We incorporate this additional information into the Bayesian model. Furthermore, we show how to select prior distributions and parameters, a task which is specific to anomaly detection. Because the ratio of outliers to inliers is so skewed, this step is crucial to ensuring accurate output. To our knowledge, we are the only work that applies probabilistic updating to anomaly detectors. 

\section{Anomaly Detection Algorithms}
We choose several anomaly detectors as inputs to the ensemble. Our goal is not to select the optimal set of detectors, but rather to focus on finding the optimal weights of already selected detectors. For literature on selecting independent detectors, see Zimek et. al for work in this direction \cite{Zimek}. We briefly describe the input detectors below. We introduce the following notation. A time series at time $t_{-n}, t_{-n+1},...,t_{-1}, t_{0}$ has associated values $y_{-n}, y_{-n+1},...,y_{-1}, y_{0}$. A detector $k$ will output a response $j_t \in \{-1,1\}$ and a probability $p_{j,t}$ that indicates its belief of the probability that the output belongs to the predicted class. 
\subsection{Variance Model}
A simple but effective detector that classifies anomalies if the change between two points is greater than a specified threshold. Define
$$
\Delta_t = y_t - y_{t-1}
$$
And a threshold $\epsilon$. The output is simply:
\[ j_t = \begin{cases} 
      1 & \Delta_t \geq \epsilon \\
      -1 & \Delta_t < \epsilon 
   \end{cases}\]

We assume the $\Delta$s follow a normal distribution, and so the output $p_{j,t}$ is related to the cumulative distribution function of the normal distribution.
$$
p_{j,t} = \Phi({\Delta_t})
$$
\subsection{Holt-Winters Algorithm}
The multiplicative Holt-Winters algorithm decomposes a time series into a deseasonalized level, a trend, and a seasonal index using exponential smoothing. Prediction of $y_t$ is then given by 
$$
\bar{y}_t = (\bar{R}_{t-1} + \bar{G}_{t-1})\bar{S}_{t-L}
$$
where \\
$\bar{R}_{t-1}$ is the deseasonalized level, \\
$\bar{G}_{t-1}$ is the trend, and \\
$\bar{S}_{t-L}$ is the seasonal index. \\
See Kalekar for methods on estimation of these variables. \cite{Kalekar} We assume the residual, $\bar{y}_t - y_t$, follows a normal distribution, and we output the probability that the error is equal to or larger than the current error, just as in the previous section.
\[ j_t = \begin{cases} 
      1 & \Delta_t \geq \epsilon \\
      -1 & \Delta_t < \epsilon 
   \end{cases}\]
$$
p_{j,t} = \Phi(\bar{y}_t - y_t)
$$ 
\subsection{Goldilocks Algorithm}
The Goldilocks algorithm uses non-negative least squares regression to predict $y_t$. At times $t_{-n}, t_{-n+1},...,t_{-1}$ we have values $y_{-n}, y_{-n+1},...,y_{-1}$. Let $T$ be this vector of timestamps, and let $Y$ be this vector of values. We perform a non-negative least squares regression such that $||Y-W^TT||^2$ is minimized, where $W$ is the weights vector. We use the weights to make a prediction $\bar{y}_t$. As in the previous section, we assume the residual, $\bar{y}_t - y_t$, follows a normal distribution, and we output the probability that the error is equal to or larger than the current error.
\[ j_t = \begin{cases} 
      1 & \Delta_t \geq \epsilon \\
      -1 & \Delta_t < \epsilon 
   \end{cases}\]
$$
p_{j,t} = \Phi(\bar{y}_t - y_t)
$$ 
\subsection{ARMA models}
We fit an auto-regressive moving average model to the time series and forecast each point $y_t$. An $ARMA(p,q)$ model has $p$ auto-regressive terms and $q$ moving average terms. The forecast is given by
$$
\bar{y}_t = c + \varepsilon_t +  \sum_{i=1}^p \varphi_i y_{t-i} + \sum_{i=1}^q \theta_i \varepsilon_{t-i}
$$
Parameters $\varphi$ and $\varepsilon$ are fitted using exact maximum likelihood. The output is again of the same form as the previous sections. 
\[ j_t = \begin{cases} 
      1 & \Delta_t \geq \epsilon \\
      -1 & \Delta_t < \epsilon 
   \end{cases}\]
$$
p_{j,t} = \Phi(\bar{y}_t - y_t)
$$ 
\subsection{Score Normalization}
Although the output of individual detectors vary, we always convert to a probability in the $[0,1]$ range. This probability can be interpreted as the detector's belief that a point is an anomaly. Once this score is obtained, we can use it as prior knowledge in a Bayesian updating scheme. It is necessary to have scores on the same scale as to not inadvertently weight some detectors as a priori more important than others. 

It is not necessary, however, to assume a normal distribution of output scores. Without loss of generality, we have no knowledge of the distribution of scores outputted by each detector, and thus there is no method that is guaranteed to work better than simple Gaussian scaling. Choosing an arbitrary distribution has been shown to significantly improve ensemble performance in practice. See Kriegal et al. for further work and discussion of the performance benefits of Gaussian scaling \cite{Kriegal}. 

If there is prior knowledge of score distributions, using other normalization methods may be better suited. Gao and Tan evaluate using logistic sigmoid functions and mixture models to normalize scores \cite{Gao}. 

\section{Bayesian Updating Meta-Algorithm}
Following Kim and Ghahramani\cite{Kim} and Platanios et al.\cite{Platanios}, we adopt a Bayesian model for combining classifiers, and then we extend this work by taking the special case when classifiers are detectors. We have a vector of data points indexed by $[1,2,...,i,...,I]$. The $i$th point has true label $t_i \in \{0,1\}$. The true label $t_i$ is generated by a Bernoulli distribution with unknown parameters $\nu_i$. $p(t_i = j | \nu_i) = p_j$ represents the proportion of outliers vs inliers.

Each classifier $k$ has output $c_i^{(k)} \in \{0,1\}$. This output is generated by a Bernoulli distribution: $p(c_i^{(k)} | t_i = j) = \pi_{j,c_i^{(k)}}^{(k)}$. $\pi^{(k)}$ then represents the confusion matrix for each classifier. The parameters for $c^{(k)}$ and $t_i$ have priors. An entry of the confusion matrix, $\pi_{j,c_i^{(k)}}^{(k)}$, has a $Beta(\alpha_{jc},\beta_{jc})$ prior. The parameter $\nu_i$ has prior $Beta(\delta_i, \gamma_i)$.

\begin{tikzpicture}

  \node[latent]                            (c) {$c_i^{(k)}$};
  \node[latent, above=of c, xshift=-1.8cm] (p) {$\pi_{j,c_i^{(k)}}$};
  \node[const, above=of p, xshift=-.9cm] (a) {$\alpha_j$};
  \node[const, above=of p, xshift=.9cm] (b) {$\beta_j$};
  \node[latent, above=of c, xshift=1.8cm]  (t) {$t_i$};
  \node[latent, above=of t] (v) {$\nu_i$};
  \node[const, above=of v, xshift=-.9cm] (d) {$\delta_i$};  
  \node[const, above=of v, xshift=.9cm] (g) {$\gamma_i$};

  \edge {p,t} {c} ; %
  \edge {a,b} {p} ;
  \edge {d,g} {v} ;
  \edge {v} {t} ;
  \plate {ct} {(t)(c)(d)(g)(v)} {$I$} ;
  \plate {cp} {(c)(p)} {$K$} ;
  \plate {pab} {(p)(a)(b)} {$J$} ;

\end{tikzpicture}
 
 In summary,
 \[  c_i^{(k)} = \begin{cases} 
      t_i & with\ probability\ \pi_{t_i,t_i}^{(k)} \\
      1 - t_i & otherwise
   \end{cases}\]
 $$
 \pi_{j,j} \sim Beta(\alpha_j,\beta_j)
 $$
 $$
 t_i \sim Bernoulli(\nu_i)
 $$
 $$
 \nu \sim Beta(\delta_i, \gamma_i)
 $$
By assuming independence among the classifiers, we obtain the posterior distribution:
 $$
 p(\pi,t,\nu|c) \propto \prod_{i=1}^{I} \left\{ p_{t_i} \prod_{k=1}^{K} \pi_{t_i,c_i^{(k)}} \right\}p(\pi|\alpha,\beta)p(\nu|\delta,\gamma)
 $$
 where
 $$
 p(\pi | \alpha,\beta) = \prod_{k=1}^{K} \prod_{j=1}^{J} p(\pi_{j,j}^{(k)} | \alpha_j, \beta_j)
 $$
 and
 $$
 p(\nu | \delta, \gamma) = \prod_{i=1}^{I} p(\nu_i | \delta_i, \gamma_i)
 $$
 The missing variables $\pi,t,\nu$ can be inferred with Markov Chain Monte Carlo methods. We use the Metroplis-Hastings algorithm. 
 
 Here we deviate from previous works, because we can take advantage of additional information outputted by each classifier. If the classifier $k$ is a detector, then it will not only output $c$, but also a confidence score $z_i^{(k)} \in [0,1]$. This confidence score is interpreted as $p(c_i^{(k)} | t_i) = z_i^{(k)}$.
 
 Instead of using an uninformative prior for $\pi_j^{(k)}$, we can select our hyperparameters using the following method. We use techniques from classical (non-Bayesian) statistics. 
 
 Let $z_i$ be a vector of $z_i^{(k)}: [z_i^{(1)},z_i^{(2)},...,z_i^{(K)}]$. We introduce the Poisson binomial distribution, which is a sum of independent but not necessarily identical Bernouilli distributions. It has probability mass function
 $$
 P(W = w) = \sum\limits_{A\in F_w} \prod\limits_{k\in A} z_i^{(k)} \prod\limits_{l\in A^c} \left(1-z_i^{(l)}\right), 
 $$
 where  $F_w$ is the set of all subsets of size $w$ that can be selected from $\{1,2,3,...,K\}$. Let us denote $Z = P(W > \lfloor K/2 \rfloor)$. A natural point estimator is then
 $$
 p(t_i = j) = Z
 $$
Intuitively, this is the probability that a majority vote of individual detectors is correct, given their individual accuracy rates. Each entry of the confusion matrix can also be obtained with the point estimator
$$
\hat{\pi}_{j,c_i^{(k)}} = Zz_i^{(k)}
$$
Now we choose hyperparameters $\alpha_j,\beta_j$ such that 
$$
E[\pi_{j,c_i^{(k)}}] = \hat{\pi}_{j,c_i^{(k)}}
$$
and the variance of $\hat{\pi}_{j,c_i^{(k)}}$ and $Beta(\alpha_j, \beta_j)$ are the same.

In similar fashion, we have an estimator for $t_i$. 
\[ \hat{t}_i = \begin{cases} 
      1 & Z > .5 \\
      0 & otherwise 
   \end{cases}\]
We choose hyperparameters $\delta_i,\gamma_i$ such that
$$
E[\nu] = E[\hat{t_i}]
$$
and the variance of $\hat{t_i}$ and $Beta(\delta_i, \gamma_i)$ are the same.
\section{Results and Discussions}
We tested the performance of our algorithms in a real-world dataset provided by Yahoo! Research \cite{Yahoo}. The dataset consists of 67 time-series which represent portions of traffic to Yahoo web properties. In total there are more than 80,000 data points and more than 1,500 anomalies. We evaluate each detector individually, a majority voting model, and our Bayesian updating model. Each model is trained separately on each time-series; i.e., learned parameters do not carry over. The threshold for individual detectors was set at $3\sigma$, i.e., we detected an anomaly if the output score was three standard deviations above the mean.
\subsection{Accuracy}
\begin{center}
\begin{tabular}{ |c|c|c|c|c|c|c| } 
 \hline
 \ & Var & Goldi & HW & ARMA & MajVote & Bayes \\ 
 \hline
 False Neg & 1133 & 993 & 943 & 1347 & 1039 & 1084 \\ 
 \hline
 True Neg & 80819 & 78614 & 79917 & 92060 & 80933 & 92611 \\ 
 \hline
 False Pos & 1138 & 3343 & 2040 & 1137 & 1024 & 586 \\
 \hline
 True Pos & 520 & 660 & 710 & 322 & 614 & 585 \\
 \hline
 Error rate & .0272 & .0519 & .0357 & .0262 & .0247 & .0176\\
 \hline
\end{tabular}
\end{center}
The Bayesian model has a 28.7\% decrease in error rate compared to the second best model, Majority Vote. We did not include the raw BCC model described in Kim and Ghahramani. During testing, because of the use of uninformative priors, the BCC model did not detect any anomalies. 
\subsection{Robustness}
A major benefit of the Bayesian model is its insensitivity to inaccurate detectors. We introduce a new detector in the ensemble, Random Detector, which randomly outputs a 1 or 0 with a reported confidence $z$ of 1. We then compare the accuracy of the Bayesian model against the Majority Voting scheme. 
\begin{center}
\begin{tabular}{ |c|c|c| } 
 \hline
 \ & MajVote & Bayes \\ 
 \hline
 False Neg & 805 & 1084 \\ 
 \hline
 True Neg & 76998 & 92611 \\ 
 \hline
 False Pos & 4959 & 586 \\
 \hline
 True Pos & 848 & 585\\
 \hline
 Error rate & .0689 & .00176 \\
 \hline
\end{tabular}
\end{center}
Predictably, the accuracy of the majority voting ensemble drops, and its error rate increases by 179\%. However, the Bayesian model is more robust; in fact, the performance is the exact same. The ensemble has learned that the random detector is error-prone and discounted its weight. To see this, we may also want to investigate the posterior probability of the confusion matrix directly. For example, we may look at $\pi_{1,1}^{(k)}$, the true positive rate, for the first time series. A random detector has a true positive rate of $.5$. The estimated posterior is:
$$
\hat{\pi}_{1,1}^{(k)} \sim Bernoulli(.522)
$$

\begin{figure}[h]
\caption{Sampled parameter of posterior distribution, true positive rate}
\centering
\includegraphics[width=0.5\textwidth]{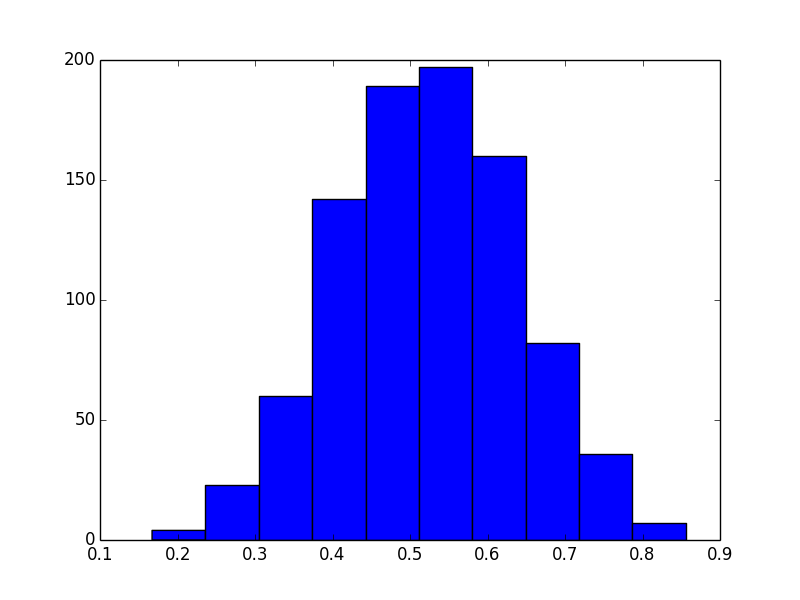}
\end{figure}

This information can be used in a variety of ways. One interesting avenue that needs to be explored further is using the confusion matrix to dynamically adjust which detectors belong in the ensemble.
\section{Conclusion}
We have shown that a Bayesian updating scheme for anomaly detection ensembles has practical as well as theoretical benefits. It provides transparency into which detectors are likely to have high error rates and allows for detailed performance analysis. The Bayesian model is also relatively insensitive to underperforming detectors. Further work can proceed in a few directions.

Some of the detectors may have correlated errors. A Markov network can be employed to detect dependence among detectors, and adjust the posterior probabilities accordingly. This was touched upon in Kim and Ghahramani and Platanios et. al \cite{Kim} \cite{Platanios}. Alternatively, rigorous methods for constructing ensembles can be explored, so that each individual detector is known to be uncorrelated with the others.

Assuming that detectors output scores which fit a normal distribution is also restrictive. More analysis of parameter fitting models for score normalization has high potential to improve the accuracy of the ensemble.

Finally, although most parameters of the model are learned, there is one that is arbitrarily set: the threshold $\epsilon$ for which we detect anomalies. As outlined in Gao and Tan, this threshold can also be learned via a Bayesian risk model, given that the penalties for false negatives and false positives are known \cite{Gao}.
\bibliographystyle{abbrv}
\bibliography{sample}


\end{document}